\documentclass[conference]{IEEEtran}
\IEEEoverridecommandlockouts
\usepackage{cite}
\usepackage{amsmath,amssymb,amsfonts}
\usepackage{algorithmic}
\usepackage{graphicx}
\usepackage{textcomp}
\usepackage{xcolor}
\def\BibTeX{{\rm B\kern-.05em{\sc i\kern-.025em b}\kern-.08em
    T\kern-.1667em\lower.7ex\hbox{E}\kern-.125emX}}

\usepackage{hyperref}
\hypersetup{
    colorlinks=true,
    linkcolor=blue,
    filecolor=magenta,      
    urlcolor=cyan,
    pdftitle={Overleaf Example},
    pdfpagemode=FullScreen,
}

\begin{document}

\title{Quasi-orthogonality and intrinsic dimensions as measures of learning and generalisation \thanks{Ivan Y. Tyukin was supported by the UKRI Turing AI Acceleration Fellowship: Adaptive, Robust, and Resilient AI Systems for the FuturE, EP/V025295/1, and by the  UK Trustworthy Autonomous Systems Verifiability Node, EP/V026801/1.}
}

\author{\IEEEauthorblockN{1\textsuperscript{st} Qinghua Zhou}
\IEEEauthorblockA{\textit{School of Computing}\\
\textit{ and Mathematical Sciences} \\
\textit{University of Leicester}\\
Leicester, UK \\
qz105@leicester.ac.uk}
\and
\IEEEauthorblockN{2\textsuperscript{nd} Alexander N. Gorban}
\IEEEauthorblockA{\textit{$\quad \quad \quad $ $\quad \quad $  School of Computing $\quad \quad \quad $ $\quad \quad $ }\\
\textit{ and Mathematical Sciences} \\
\textit{University of Leicester}\\
Leiceser, UK\\
a.n.gorban@leicester.ac.uk}
\and 
\IEEEauthorblockN{3\textsuperscript{rd} Evgeny M. Mirkes}
\IEEEauthorblockA{\textit{School of Computing }\\
\textit{ and Mathematical Sciences} \\
\textit{University of Leicester}\\
Leiceser, UK\\
em322@leicester.ac.uk}
\and
\IEEEauthorblockN{4\textsuperscript{th} Jonathan Bac}
\IEEEauthorblockA{\textit{Institut Curie} \\
\textit{$\quad \quad \quad $ PSL Research University $\quad \quad \quad $}\\
Paris, France \\
jonathan.bac@cri-paris.org}
\and
\IEEEauthorblockN{5\textsuperscript{th} Andrei Zinonyev}
\IEEEauthorblockA{\textit{Institut Curie} \\
\textit{$\quad \quad \quad \quad \quad$ PSL Research University $\quad \quad \quad \quad \quad$}\\
Paris, France \\
andrei.zinovyev.u900@gmail.com}
\and
\IEEEauthorblockN{6\textsuperscript{th} Ivan Y. Tyukin}
\IEEEauthorblockA{\textit{Department of Mathematics} \\
\textit{King's College London}\\
London, UK \\
ivan.tyukin@kcl.ac.uk}
}

\maketitle

\begin{abstract}
Finding best architectures of learning machines, such as deep neural networks, is a well-known technical and theoretical challenge. Recent work by Mellor et al \cite{mellor2021neural} showed that there may exist correlations between the accuracies of trained networks and the values of some easily computable measures defined on randomly initialised networks which may enable to search tens of thousands of neural architectures without training. Mellor et al \cite{mellor2021neural} used the Hamming distance evaluated over all ReLU neurons as such a measure. Motivated by these findings, in our work, we ask the question of the existence of other and perhaps more principled measures which could be used as determinants of success of a given neural architecture. In particular, we examine, if the dimensionality and quasi-orthogonality of neural networks' feature space could be correlated with the network's performance after training.  We showed, using the setup as in Mellor et al \cite{mellor2021neural}, that dimensionality and quasi-orthogonality may jointly serve as network’s performance discriminants. In addition to offering new opportunities to accelerate neural architecture search, our findings suggest important relationships between the networks’ final performance and properties of their randomly initialised feature spaces: data dimension and quasi-orthogonality.
\end{abstract}


\section{Introduction}

The advent of deep neural networks in computer vision has provided significant strides in various benchmarks and practical applications. A key problem in every application and practical experimentation is the design of neural network architecture. Neural architecture search (NAS) is a field in deep learning where the design task has shifted from manual selection to algorithmic optimisation. 

The number of different neural architectures is vast. In general, even if a task at hand is given and known, it is not clear if one can fully explore the set of all practically implementable neural architectures for that task. A more practical approach would be to fix a reasonably large relevant class of architectures and search for the best architecture in that class. However, despite that this significant relaxation of the problem may potentially offer hope, the task of exploring finite yet large classes of architectures remains a major technical and computational challenge. To put this into perspective, if one is to explore $10^4$ architectures, each requiring one hour to train, then an experiment involving the assessment of $10$ training runs per architecture would take more than $11$ calendar years to complete. 

A very recent work \cite{mellor2021neural} proposed an intriguing and apparently counter-intuitive approach to the NAS challenge. Instead of an exhaustive trial-and-error analysis of the final performance of trained models, the authors suggested to look at the correlations between the performance of trained models and properties of data representations, such as the Hamming distance between neural activation functions, in latent spaces of untrained networks. Surprisingly, empirical evidence reported in \cite{mellor2021neural} suggests that this approach may indeed be practical.

Inspired by these empirical results, in this paper we explore the feasibility of using other characterisations of data representations in untrained networks for NAS. In particular, we explore how various measures of data intrinsic dimensionality and quasi-orthogonality in the networks' feature spaces may correlate with the final performance of trained networks. In our explorations we considered $9$ different notions and measures of data dimension: Correlation Dimension \cite{grassberger2004measuring}, Fisher Separability dimension \cite{albergante2019estimating}, k-Nearest Neighbours dimension \cite{carter2009local}, Principal Component Analysis - based dimension \cite{fukunaga1971algorithm}, Manifold-Adaptive Dimension  \cite{farahmand2007manifold}, MiND\_MLk and MiND\_MLi \cite{rozza2012novel},
Maximum Likelihood \cite{haro2008translated}, Method Of Moments \cite{amsaleg2018extreme}, TwoNN dimension \cite{facco2017estimating}. To compute these dimensions for a given data sample we used scikit-dimension package \cite{bac2021scikit}.

In contrast to the conventional notion of dimension as the minimal number of linearly independent vectors spanning the set, notions of data intrinsic dimension such as Fisher Separability dimension, capture fine-grain structure of concentrations relative to elementary volumes of the ambient space. Similar properties have been studied recently in the context of understanding generalisation in overparameterised AI models \cite{bartlett2020benign}. 

Our study shows that, for the architectures used in NAS-bench201, data intrinsic dimension, measured by Fisher Separability dimension of data batches in models' latent spaces, could serve as a viable attribute for selecting neural architectures whose trained performance is at the top of the range. Strikingly, we see that the smaller the value of Fisher Separability dimension, the lower is the risk of picking an under-performing architecture. On the other hand, data orthogonality, which was shown to be particularly relevant for learning from few examples in \cite{tyukin2021demystification}, demonstrated a somewhat reciprocal correlation with the trained networks' performance: lower orthogonality correlated with high final performance.

The paper is organised as follows. In Section \ref{sec:preliminaries} we present  background information explaining our NAS experimental settings, and provide relevant details on quasi-orthogonality and data intrinsic dimenisonality used to quantify and compute these quantities in experiments. Section \ref{sec:results} contains main results of the study, including experimental findings and discussion. Section \ref{sec:conclusion} concludes the work.

\section{Preliminaries}\label{sec:preliminaries}

\subsection{NASbench201 feature space}

Neural architectural search benchmarks are constructed to provide a defined search space for evaluation and comparison between NAS algorithms. The benchmark contains a number of predefined architectures and corresponding metadata, including the loss and accuracy of individual architectures. These metrics indicate the learning and generalisation of said architectures if trained under uniform conditions. In the following text, we will refer to these metadata as metrics. As inspired by Mellor et al. \cite{mellor2021neural}, measures can be identified from untrained architectures as determinants of potential performance.

In our work, we focused on the 50\% of all architectures within the NAS-Bench-201 benchmark \cite{dong2001bench} search space and the CIFAR-10 dataset. This combination of search space and dataset can indicate measure quality within limited computation cost. 

A key factor to consider is the training strategy of said metrics. For metrics calculated for architectures within NAS-Benchmark-201, a training strategy from \cite{dong2001bench} are stated below,

\begin{itemize}
\item SGD optimizer with Nesterov enabled (and momentum factor of 0.9) \& weight decay of $1\times10^{-5}$
\item Categorical cross-entropy loss function
\item Batch size of 256 samples
\item Initial learning rate (LR) of 0.1 with cosine learning rate schedule
\item Maximum epoch of 200 regardless of learning and generalisation
\item Number of times to stack cells $N=5$
\item Nodes within the densely-connected directed acyclic graph to represent cells $V=4$
\item Initial channels of 16 and initial batch normalisation layer
\end{itemize}

The above strategy indicates that learning and generalisation are not guaranteed for individual architectures. Therefore the correlation between measures and metrics cannot be assumed to be independent of the above strategy, and the following statements of learning and generalisation are based within this context. With the practical limitations of training in real-world scenarios, it is unlikely we can optimise to the full extent of any architecture; therefore, this work can still provide useful insight into factors affecting learning and generalisation, including quasi-orthogonality and data intrinsic dimensionality.

\subsection{Measures of Quasi-orthogonality}

Quasi-orthogonality is a property where data samples become almost orthogonal to each other in high dimensional feature space. This property is an implication of the "blessing of dimensionality", and an underlying reason for high linear separability for high dimensional data \cite{gorban2018blessing}. Previous work has proven the "typicality" of quasi-orthogonality if distributions of data in the high-dimensional latent space are supported on $n$-spheres and satisfy non-degeneracy constraints \cite{tyukin2021demystification}. In this work, we found basic orthogonality measures based on the previous theoretical foundation, including 

\begin{itemize}
    \item Mean and standard deviation of angles between individual samples (\texttt{f\_mean} and \texttt{f\_std})
    \item Mean and standard deviation of angles between samples and their corresponding class centroids (\texttt{cmean} and \texttt{cstd})
\end{itemize}

For each network, features extracted for the 1280 samples were first centred via its mean as a preprocessing step prior to calculating class centroids and angles between samples.

\subsection{Measures of Intrinsic Dimensionality}

In machine learning, methods for the estimation of data intrinsic dimensionality are diverse. The criteria and underlying principles for determining the effective number of variables needed to approximate the data vary between methods. 

A range of data intrinsic dimensions measures were calculated with the \texttt{scikit-dimension} package\footnote{https://github.com/j-bac/scikit-dimension} on the selected feature space of each network. Intrinsic dimensionality measures include estimations with:
\begin{itemize}
    \item Correlation Dimension - \texttt{CorrInt} \cite{grassberger2004measuring}
    \item Fisher Separability algorithm - \texttt{FisherS} \cite{albergante2019estimating}
    \item k-Nearest Neighbours algorithm - \texttt{KNN} \cite{carter2009local}
    \item Principal Component Analysis - \texttt{lPCA} \cite{fukunaga1971algorithm}
    \item Manifold-Adaptive Dimension Estimation algorithm - \texttt{MADA} \cite{farahmand2007manifold}
    \item MiND\_MLk and MiND\_MLi algorithms - \texttt{MIND\_ML} \cite{rozza2012novel}
    \item Maximum Likelihood algorithm - \texttt{MLE} \cite{haro2008translated}
    \item Method Of Moments algorithm - \texttt{MOM} \cite{amsaleg2018extreme}
    \item TwoNN algorithm - \texttt{TwoNN} \cite{facco2017estimating}
\end{itemize}

These measures have diverse underlying theories to dimensionality. Based on empirical observations, we place a particular focus on the Fisher Separability algorithm (\texttt{FisherS})  \cite{albergante2019estimating}. The Fisher separability algorithm requires preprocessing the data $X$ via mean centring, PCA dimensionality reduction, whitening and projection to the unit sphere. Fisher linear-separability between $X$ and a point cloud $Y$ is defined as $(x,y)\leq\alpha(x,x)$, $\forall y\in Y, x\in X$ and $\alpha\in[0,1)$. The  measure of Fisher separability is quantified as,
\begin{equation}
    \text{ID}_{\texttt{FisherS}} = \frac{W\left(\frac{-\ln(1-\alpha^2)}{2\pi\bar{p}^2_\alpha \alpha^2(1-\alpha^2)}\right)}{-\ln{(1-\alpha^2)}}
\end{equation}
where $\bar{p}_\alpha=\left((1-\alpha^2)^{(n-1)/2}\right)/\left(\alpha\sqrt{2\pi n}\right)$ is the mean probability of individual points been separable from all other points, $W$ is the Lambert function. In practice, a range of $\alpha$ is profiled and an ad hoc estimate is taken for $\alpha=0.8\max\left\{\alpha_m | \bar{p}_\alpha >0 \right\}$. The same $n$-sphere assumption as quasi-orthogonality is applied in this scenario. For more detailed algorithm and procedures, please refer to the original work by Albergante et al \cite{albergante2019estimating}.

\begin{figure}
\centerline{\includegraphics[width=1\columnwidth]{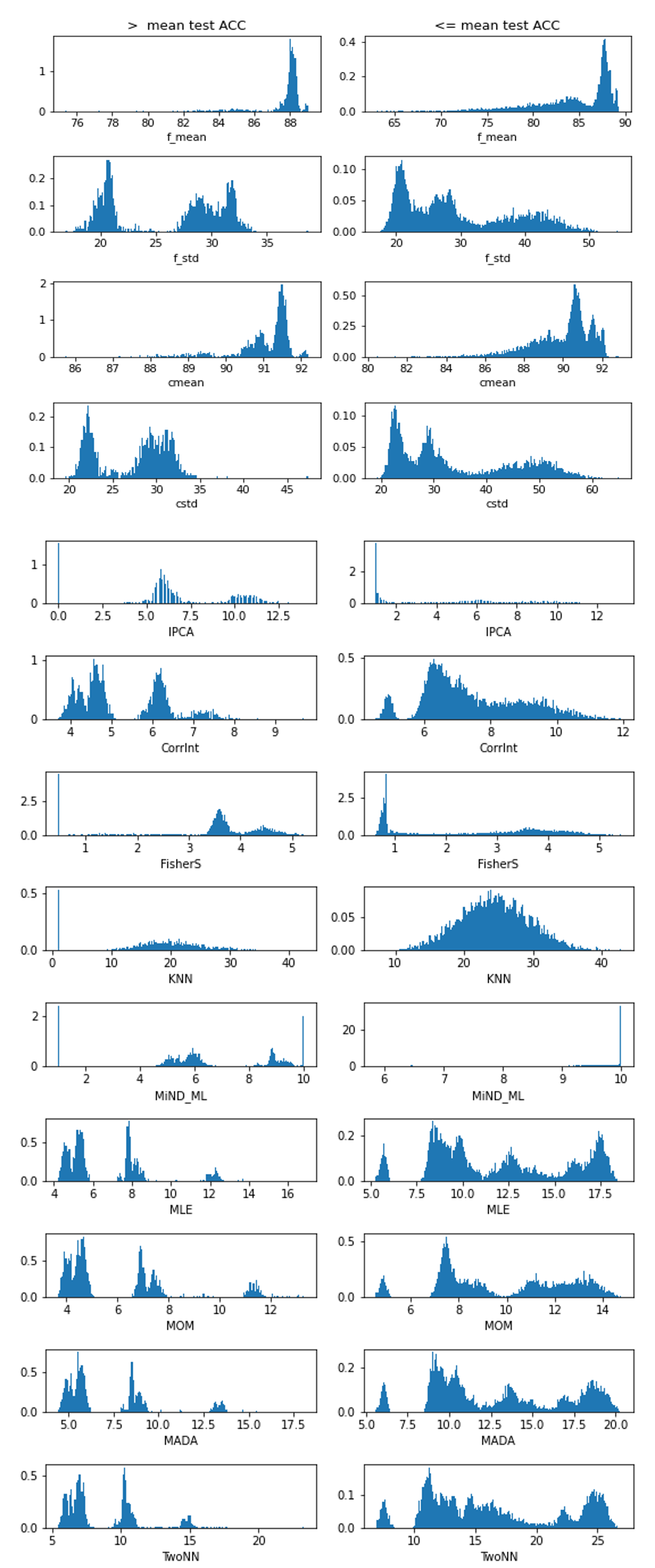}}
\caption{Density histograms of architectures with (1) test accuracy $>$ mean test accuracy, and (2) test accuracy $\leq$ mean test accuracy of the sampled NASbench201 search space. The top four histograms correspond to the distributions of orthogonality measures, while the remainder corresponds to the distributions of data intrinsic dimensionality measures.}
\label{fig1}
\end{figure}

\section{Results}\label{sec:results}

Features were extracted from the global average pooling activation before the fully connected classification layer for a single feature space. Properties of the NAS-Bench-201 search space indicates that the feature space will be of the same  ($n=64$) for all architectures. Measures of quasi-orthogonality and data intrinsic dimensionality are derived directly from the feature space of each architecture. For valid measures of data intrinsic dimensionality, we used ten random batches of 128 samples from the CIFAR-10 dataset as input for feature extraction. 

Since weights of the initial weights used by the original benchmark study \cite{dong2001bench} cannot be reproduced with confidence, we sampled weights from the Kaiming initialisation, i.e. $U(-b,b)$, for $b = g \sqrt{3/D_i}$, where $g$ is a scaling factor and $D_i$ is the layer input dimension. The sampling was performed 50 times, and we extracted features for each initialisation. We calculated orthogonality and data intrinsic dimensionality measures for each sampling, and the average was taken as the corresponding measure for the architecture. The standard deviations for all measures apart from the data intrinsic dimensionality measure of  \texttt{KNN} have been observed to be small.

The main metric of choice is test accuracy; this metric is extracted from the NASbench201 metadata and based on the test accuracy of architecture if it is trained via conditions listed in Section II. The following results and figures are based on the test accuracy for more direct relation with generalisation. This choice is based upon the observation that the same qualitative results have been observed via validation accuracy.

\subsection{Empirical Quasi-orthogonality correlations with NASbench201 performance}

Qualitative observations can be made via examining the distribution of orthogonality measures for networks performance. In Figure \ref{fig1}, the first four rows present four pairs of histograms of orthogonality measures for networks with test accuracy greater than, or smaller and equal to the mean test accuracy of the sampled NASbench201 search space. Evident differences in distribution between the two histograms of each pair can be observed for all four orthogonality measures. These observations provide the basis for selecting quasi-orthogonality as an indicator of network performance.

\begin{figure}[h]
\centerline{\includegraphics[width=1\columnwidth]{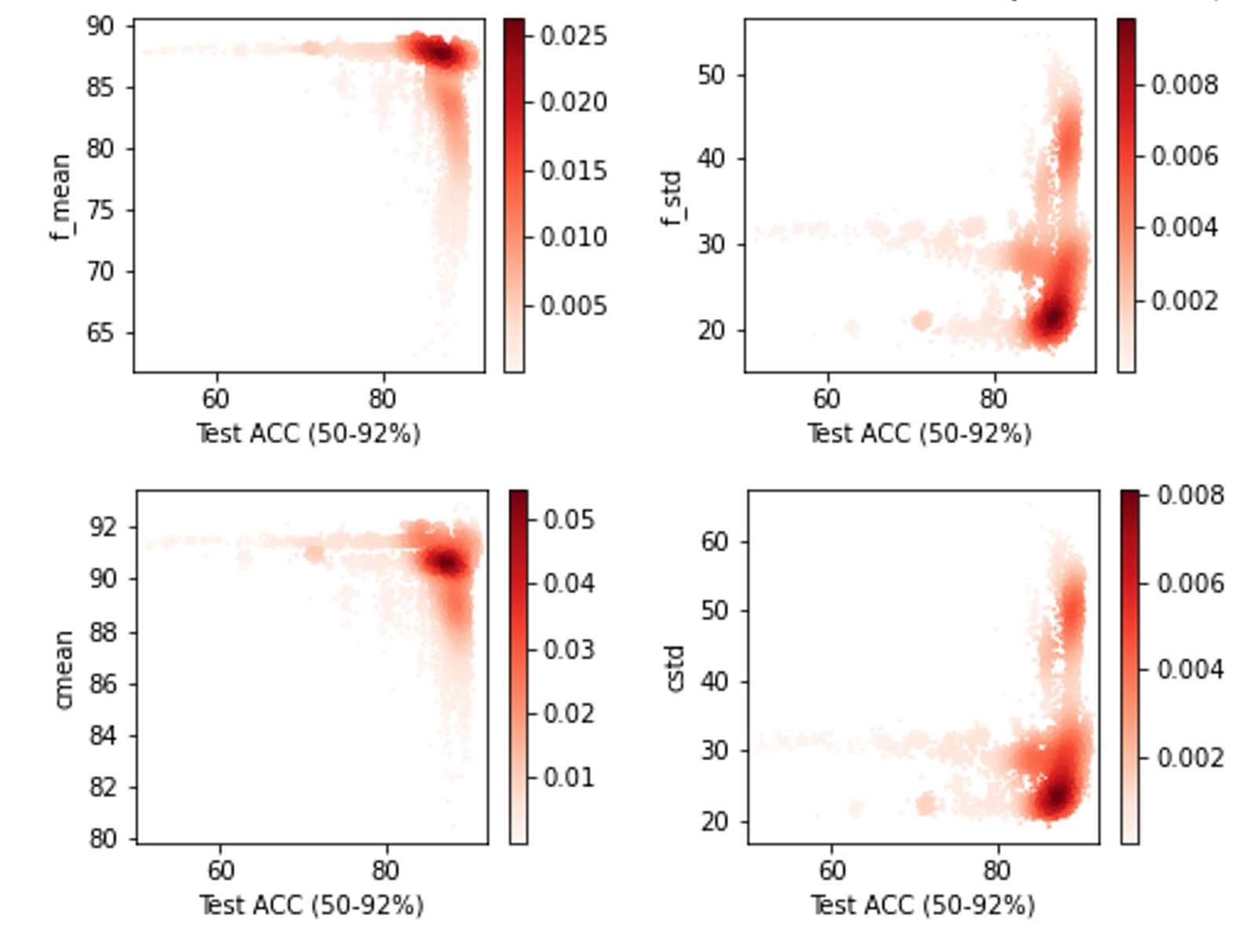}}
\caption{Scatter plots between orthogonality measures and test accuracy (with focus on 50-92\% accuracy region). Heat-map reflecting the quantity of architectures on the scatter plot calculated via Gaussian kernel-density estimate.}
\label{fig2}
\end{figure}

We also provide scatter plots between orthogonality measures (averaged over multiple weight sampling) and the test accuracy of the networks if trained (Figure \ref{fig2}). A clear concentration of high performing architectures in the high mean orthogonality and low standard deviation can be observed. The empirical results presented in this work also have further validated the theoretical proof for the "typicality" of quasi-orthogonality. This theory provides the foundation for our understanding of few-short, and one-shot learning \cite{tyukin2021demystification}.

Though no linear relationship between the measures and accuracy can be observed, it is evident there exists a reciprocal correlation. Therefore, lower orthogonality (\texttt{f\_mean} and \texttt{cmean}) are potential indicators for avoiding low performing architectures. These architectures are observed as a horizontal "tail" in the 80-90 degrees range for \texttt{f\_mean} and 88-92 degrees range for \texttt{cmean}. Similar observations can be made for standard deviations (STD) of orthogonality, where high STD (\texttt{f\_std} and \texttt{cstd}) are similar indicators for avoiding low performing architectures. In addition, a smaller "tail" is observed for STD of around 20 degrees compared to 30 degrees, while a high concentration of architectures exists around 20 degrees and the quasi-orthogonal region.

\subsection{Empirical Data dimensionality correlations with NASbench201 performance}

\begin{figure}[h]
\centerline{\includegraphics[width=1\columnwidth]{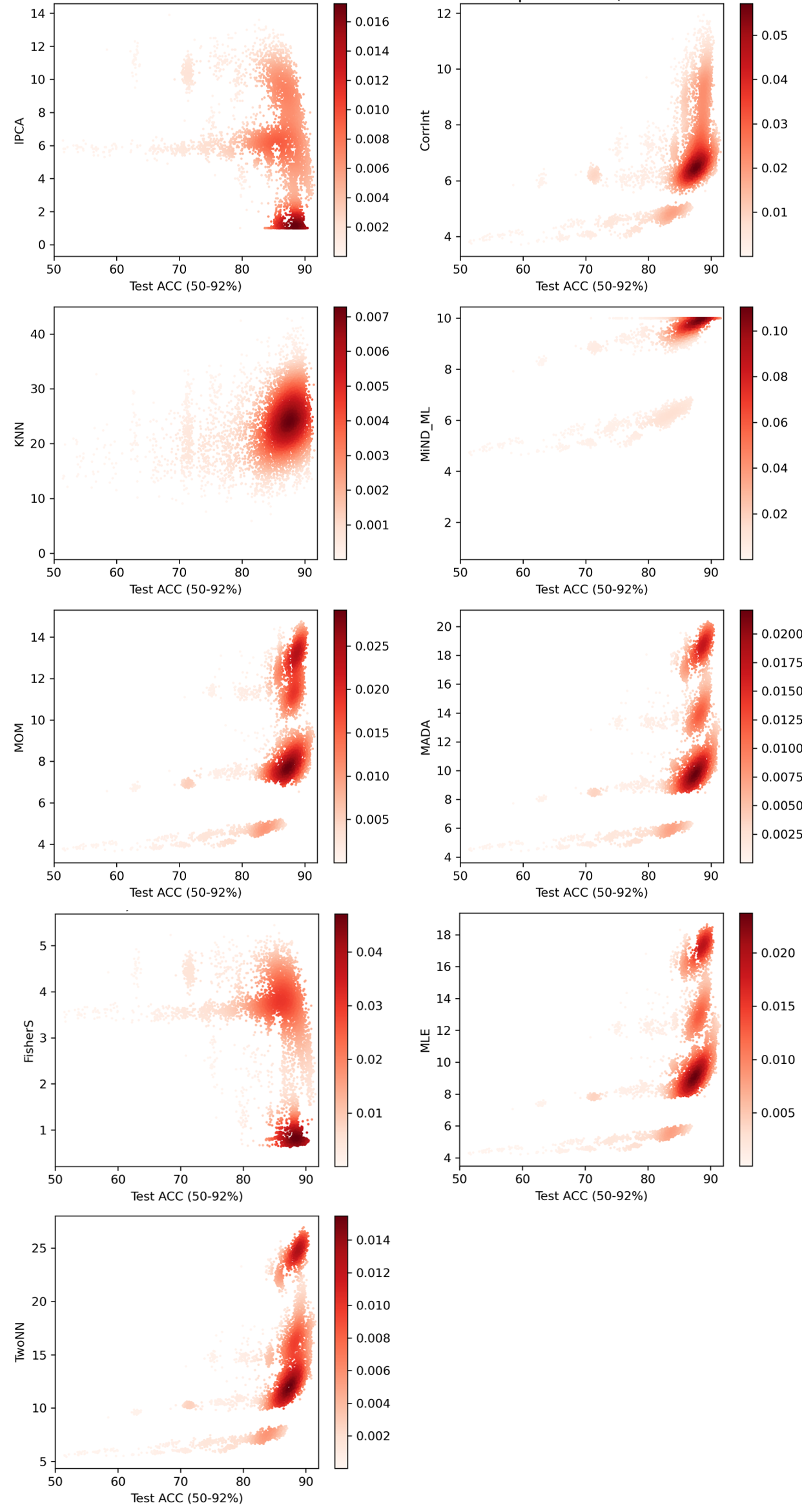}}
\caption{Scatter plots between data intrinsic dimensionality measures and test accuracy (with focus on 50-92\% accuracy region). Heat-map reflecting the quantity of architectures on the scatter plot calculated via Gaussian kernel-density estimate.}
\label{fig3}
\end{figure}

The same histograms pairs based on binary separation of architectures via the mean test accuracy can be found in rows 5-13 in Figure \ref{fig1}. Similar to the orthogonality measures, evident differences in distribution can be found between the distribution of the two groups of architectures. Figure \ref{fig3} contains scatter-plots between data intrinsic dimensionality measures and test accuracy within the sampled NASbench201 search space. Similar to orthogonality measures, these empirical observations reveal correlations between multiple ID measures to architecture's performance in NASbench201.

A range of interesting observations can be made based on these results, including the concentration of high performing architectures in the low intrinsic dimension region ($\text{ID}=1,2$) for \texttt{lPCA} and \texttt{FisherS}. This observations provides a simple limit can avoid selection within spread of relatively lower performing architectures in the $\text{ID}>=3$ region. Similar observations can be made for the $\text{ID}_{\texttt{MIND\_ML}}\leq8$, $\text{ID}_{\texttt{CorrInt}}\leq5$, $\text{ID}_{\texttt{MLE}}\leq6$, $\text{ID}_{\texttt{MOM}}\leq6$, $\text{ID}_{\texttt{MADA}}\leq6$ and $\text{ID}_{\texttt{TwoNN}}\leq8$ regions. Filtering out architectures within theses regions of ID measures can potentially aid in avoiding lower performing architectures.

In addition, for measures of \texttt{MLE}, \texttt{MOM}, \texttt{MADA} and \texttt{TwoNN}, multiple horizontal "tails" can be observed for different ranges of ID measures. A commonality between these "tails" is that  (1) they show signs of linear relationships between ID measure and test accuracy, and  (2) their spread decreases with higher ID measures. These observations indicate that choosing networks with higher measures in these ID can potentially provide relatively high performing architectures.

\subsection{Discussion}

\begin{figure*}
\centerline{\includegraphics[width=2.2\columnwidth]{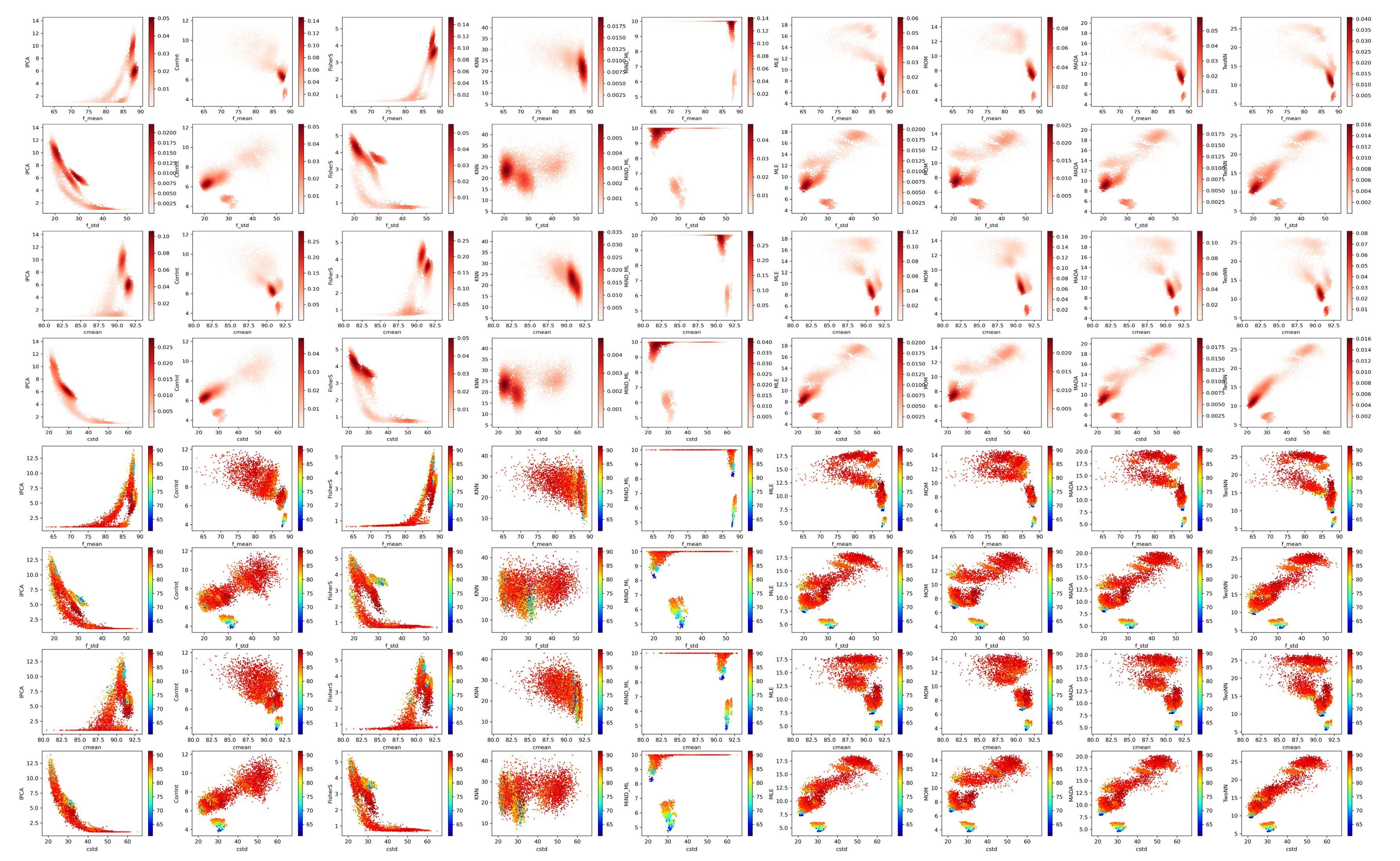}}
\caption{2D scatter plots between orthogonality measures and data intrinsic dimensionality measures. The heat maps of the top four rows of scatter plots are based on the quantity of architectures calculated via Gaussian kernel-density estimate. The heat maps of the bottom four rows of scatter plots are based on test accuracy (within the 60-93\% range).}
\label{fig4}
\end{figure*}

The previous sections show that orthogonality and data intrinsic dimensionality (ID) measures can act as computationally efficient determinants that correlate with performance. However, there exist limitations to both types of measures. Regions of choice preferable for ID measures shows high concentrations of architectures, e.g. $\text{ID}_{\texttt{FisherS}}\leq2$.  However, the most highly performing architectures (extreme points) within are not contained for both orthogonality and ID measures. Therefore, we explore the combination of the two types of measures. Figure \ref{fig4} presents two sets of scatter plots between the two types of measures with heat maps indicating (1) quantity of architectures and (2) test accuracy.

\begin{figure*}
\centerline{\includegraphics[width=2.2\columnwidth]{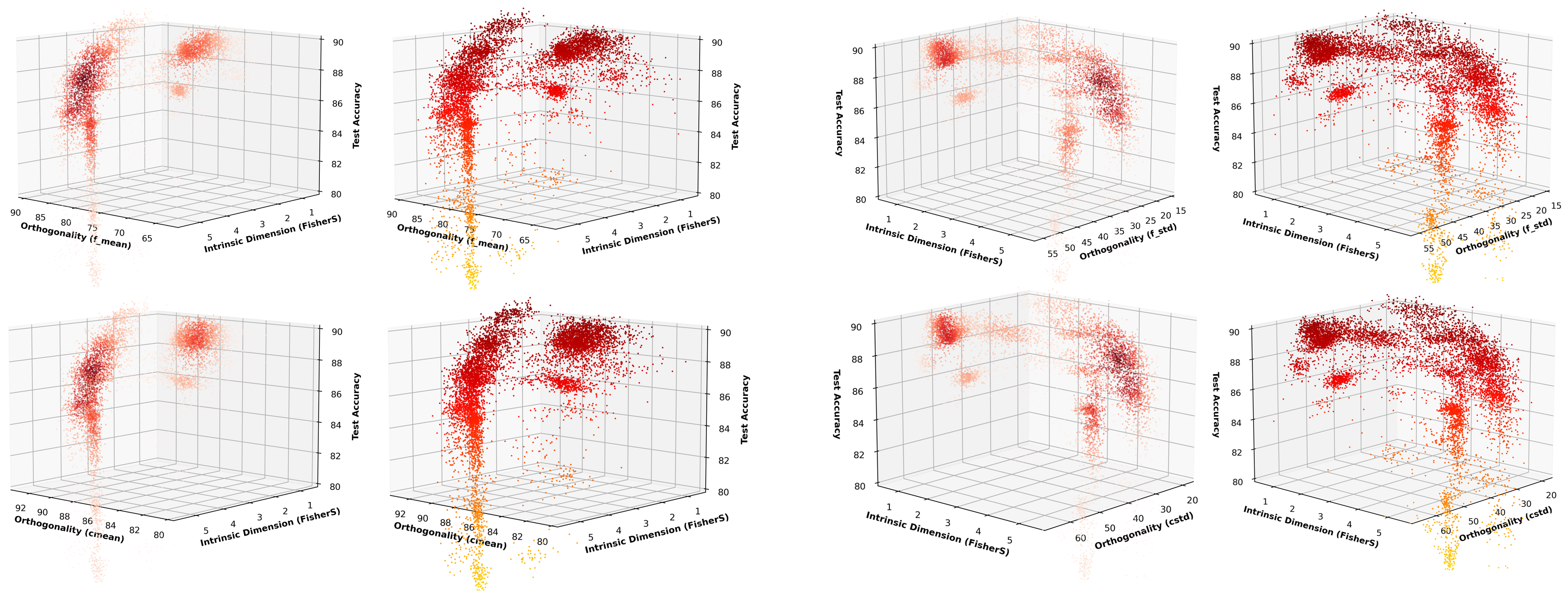}}
\caption{3D scatter plots between orthogonality measures, Fisher separability measure of data intrinsic dimensionality, and test accuracy.}
\label{fig5}
\end{figure*}

Observations can be made on the plots between \texttt{FisherS} and all orthogonality measures, where the higher orthogonality measures correlate to higher \texttt{FisherS}. To provide a better qualitative view of our results, we present 3D scatter plots between orthogonality measure, ID measure of \texttt{FisherS} and test accuracy in Figure \ref{fig5}. From the figure, we can see a possible separation of architectures into two main clusters and the concentration of the worst-performing architectures on one cluster (showing a long vertical "tail", which we will refer to via quantity as the "large" cluster). This observation is consistent with prior plots of orthogonality and ID measures. Simple avoidance of this "large" cluster via selecting architectures with low $\text{ID}_{\text{FisherS}}$ and low degree of quasi-orthogonality may allow us to avoid the risk of picking a low performing architecture.

An alternative is to aim for top-performing networks. From the 2D scatter plots, a small "tail" can be be found for $\text{ID}_{\text{FisherS}}\in[1.5,2.5]$ and orthogonality measures \texttt{f\_mean} and \texttt{cmean} near 90 degrees, where a sparse cluster of the best performing architecture exists. Similar observations can be made for \texttt{f\_std} and \texttt{cstd}. This observation from the 2D scatter plots are more evident on the 3D plots, where a small cluster exists at the higher performance end of the "large" cluster, showing a sparse bulge of top performing architectures. Therefore these criteria ($\text{ID}_{\text{FisherS}}\in[1.5,2.5]$ with $85^{\circ}\leq\texttt{f\_mean}\leq88^{\circ}$ or $90^{\circ}\leq\texttt{f\_mean}\leq92.5^{\circ}$) can potentially aid in identifying higher performing architectures.

\section{Conclusion}\label{sec:conclusion}

The design of deep neural network architectures for better learning and generalisation is a fundamental challenge in deep learning. Even with the most recent NAS algorithms, this challenge is often expensive to resolve. 

Motivated by the recent work by Mellor et al. \cite{mellor2021neural}, we utilised the neural architectural search space and metadata of NASbench201 to find principled measures of quasi-orthogonality and data intrinsic dimensionality as potential measures of learning and generalisation. 

Empirical results validate theoretical assumptions of the "typicality" of quasi-orthogonality. Data representation in feature spaces in the majority of untrained neural architectures demonstrated marked quasi-orthogonality patterns. At the same time, untrained networks with a relatively low degree of data quasi-orthogonality showed good performance after training. Trained performance for networks with high degree of initial quasi-orthogonality showed significant spread.

Independently of quasi-orthogonality, a number of the proposed measures can aid in avoiding lower-performing architectures, e.g. via selecting architectures with low \texttt{f\_mean} and \texttt{FisherS}. This can be further improved via the use of measures of orthogonality with intrinsic dimensionality $\text{ID}_\texttt{FisherS}$. Notably, we observed a sizable fraction of networks whose latent representations shared both, low Fisher Separability dimension and a high degree of quasi-orthogonality. Their trained accuracy was typically high.  

In addition, these combinations can potentially guide selection for a sparse group of top-performing architectures. These empirical findings can potentially be utilised to select architectures prior to training or filter out groups of architectures within NAS search space and reduce computational cost. This work is limited by the selected dataset and the fundamental limitations of the NAS benchmark training strategy. However, we believe this will be an additional step towards no-training or limited training NAS and further discoveries for measures governing learning and generalisation.

\bibliographystyle{IEEEtran}
\bibliography{IEEEabrv,references}

\end{document}